\documentclass{article}
\usepackage{flushend}
\usepackage{spconf,amsmath,graphicx}
\usepackage{subfigure}
\usepackage{multirow}
\usepackage{longtable}
\usepackage{amsthm}
\usepackage{amssymb }
\usepackage{amsmath}
\usepackage{bm}
\usepackage{mathrsfs}
\usepackage{amsfonts}
\usepackage{longtable}
\usepackage{multirow}
\usepackage{algorithm, algpseudocode}
\usepackage{cite}
\usepackage{color}

\def\s{\bm{s}}
\def\u{\bm{u}}
\def\v{\bm{v}}

\def\W{\bm{W}}

\algnewcommand\Input{\item[\hspace{6pt}\textbf{Input:}]}
\algnewcommand\Output{\item[\hspace{6pt}\textbf{Output:}]}
\algnewcommand\OutputVal{\textbf{output} }
\usepackage{tabularx}
    \newcolumntype{L}{>{\raggedright\arraybackslash}X}

\title{Brain Tumor Type Classification via Capsule Networks}
\name{Parnian Afshar$^\dagger$,  Arash Mohammadi$^\dagger$, and Konstantinos N. Plataniotis$^\ddagger$}
\address{$~^\dagger$Concordia Institute for Information Systems Engineering,  Concordia University, Montreal, QC, Canada \\
$~^\ddagger$Department of Electrical and Computer Engineering, University of Toronto, Toronto, ON, Canada\\
Emails: $\{$p\underline{\space}afs, arashmoh$\}$@encs.concordia.ca; kostas@ece.utoronto.ca
 \thanks{ This work was partially supported by the Natural Sciences and Engineering Research Council (NSERC) of Canada through the NSERC Discovery Grant RGPIN-2016-049988.}}
\begin{document}
\ninept
\maketitle
\begin{abstract}
Brain tumor is considered as one of the deadliest and most common form of cancer both in children and in adults.  Consequently, determining the correct type of brain tumor in early stages is of significant importance to devise a precise treatment plan and predict patient's response to the adopted treatment. In this regard, there has been a recent surge of interest in designing Convolutional Neural Networks (CNNs) for the problem of brain tumor type classification. However, CNNs typically require large amount of training data and can not properly handle input transformations. Capsule networks (referred to as CapsNets) are brand new machine learning architectures proposed very recently to overcome these shortcomings of CNNs, and posed to revolutionize deep learning solutions. Of particular interest to this work is that Capsule networks are robust to rotation and affine transformation, and require far less training data, which is the case for processing medical image datasets including brain Magnetic Resonance Imaging (MRI) images. In this paper, we focus to achieve the following four objectives: (i) Adopt and incorporate CapsNets for the problem of brain tumor classification to design an improved architecture which maximizes the accuracy of the classification problem at hand; (ii) Investigate the over-fitting problem of CapsNets based on a real set of MRI images; (iii) Explore whether or not CapsNets are capable of providing  better fit for the whole brain images or just the segmented tumor, and; (iv) Develop a visualization paradigm for the output of the CapsNet to better explain the learned features. Our results show that the proposed approach can successfully overcome CNNs for the brain tumor classification problem.
\end{abstract}
\section{Introduction} \label{sec:Introduction}
According to 2016 cancer statistics~\cite{Siegel:2016}, brain tumor is considered as the leading cause of cancer-related morbidity, and mortality around the world and is known as one of  the most common form of cancers both in children and in adults. Medical image processing~\cite{Zhang:2016} is widely used for early detection of brain cancer, which consequently results in devising more effective treatments and increases cancer survival rate. However, brain tumors have different variaties/types (e.g., Meningioma, Pituitary, and Glioma~\cite{National}), which makes determining the correct type in early stages a significantly challenging problem, but a crucial task since it helps physicians to have a more precise treatment plan and have a better prediction of patient's response to the adopted treatment. On the other hand, tumor type classification by human inspection is an extremely time consuming and error prone task, which highly depends on the experience and skills of the radiologist. This fact has resulted in a recent surge of interest~\cite{Usman:2017, Abbadi:2017, Kaur:2017, Anitha:2016, Cheng:2015, Mohsen:2017} in designing highly accurate automated image processing systems for brain tumor classification. Among different available medical imaging technologies, Magnetic Resonance Imaging (MRI) is more favored for brain tumor type classification due to its harmless nature and is also the focus of this paper.

Generally speaking, cancer tumor classification task consists of a segmenting step~\cite{Ravi:2017, Kostas:2017, Ana:2013, Roy:2013} to identify the tumor region from medical images followed by a feature extraction step, which selects useful features to be used for tumor classification~\cite{Li:2015}. Havaei \textit{et al.}~\cite{Havaei:2017} have provided one of the most recent works for brain tumor segmentation. This work has proposed a two path Convolutional Neural Network (CNN) which not only takes the pixel properties into account, but also considers the probabilities of neighboring pixels. Once the tumor region is segmented, different types of features can be extracted to be fed to the classification module, e.g.,  Usman~\textit{et al.}~\cite{Usman:2017} have used intensity, intensity differences, neighborhood, and wavelet texture as the input feature vector to train a random forest classifier. Reference~\cite{Cheng:2015} studied effect of tumor region augmentation on three feature extraction methods, i.e., intensity histogram, Gray Level Co-occurrence Matrix (GLCM), and bag-of-words (BoW). It is shown that tumor region augmentation can enhance the accuracy of brain tumor classification. Abbadi~\textit {et al.}~\cite{Abbadi:2017} have also adopted the GLCM and Gray Level Run Length Matrices (GLRLM) to extract 18 features for tumor classification using Probabilistic Neural Networks (PNNs). All the aforementioned studies on tumor classification have one considerable drawback, i.e., they need a prior knowledge regarding the type of features to be extracted, which reduces their generalization capability.

The CNNs~\cite{Alex:2012} have extensive learning capacity and can infer the nature of an input image without any prior knowledge, which makes them a suitable method for image classification. Utilization of CNNs for brain tumor type classification is recently explored in Reference~\cite{Justin:2017}, where neural networks and CNNs are used together with different pre-processing steps including data augmentation. It was shown that CNNs without any pre-processing outperform other methods on axial brain MR images. Although CNNs have successfully overcome many approaches in image processing, they still have some drawbacks. For instance, they are not robust to affine transformation and do not take the spatial relationships within the image into considerations. To improve their generalization, therefore, CNNs need to be provided with training data consisting of all kinds of rotations and transformation. Besides, CNNs typically perform poorly confronting small data sets, which is the case for most of the medical image databases, including brain MRIs.

To overcome the aforementioned drawbacks of CNNs, Sabour and Hinton~\textit{et al.} have recently proposed Capsule networks (CapsNets)~\cite{Hinton:2017} with each Capsule within the network consisting of several neurons. Activity vector of each  capsule is composed of several pose parameters (e.g., position, orientation, scaling, and skewness). The length of each activity vector provides the existence probability of the specific object represented by that Capsule. The most important property of CapsNets is called routing by agreement, which means Capsules in lower levels predict the outcome of Capsules in higher levels, and the higher level Capsules get activated only if these predictions agree.
In this paper, we capitalize on achievability of these benefits and adopt the CapsNet architecture for brain tumor type classification problem. In this regard, we explore different potential architectures of Capsule networks and identify the one which maximizes the prediction accuracy for the problem at hand. Furthermore, we consider the following two scenarios as the input to the designed CapsNet to investigate effects of input data on CapsNets: (i) The whole brain image is fed into the network, and; (ii) The segmented tumor regions are incorporated. Since Capsule networks have complicated structures and several parameters to be learned, they tend to over-fit the training data especially on small datasets like brain MRIs. For this reason, a regularization criteria is adopted to address the overfitting problem of CapsNets for brain tumor classification problem. Finally, we develop a visualization paradigm for the output of the CapsNet to better explain the features that the designed model has learned from the input.

The rest of this paper is organized as follows: Section~\ref{sec:framework} describes required mathematical background for CNNs. Section~\ref{sec:WTE}, considers CapsNets and presents the proposed approach followed by experimental results in Section~\ref{sec:EXP}. Finally, Section~\ref{sec:con} concludes the paper.

\section{Problem Formulation} \label{sec:framework}
Given a set of $N_{\text{train}}$ training MRI brain images, the goal is to design a deep learning architecture to identify/classify the type of brain tumor available in the given test MRI images into three different categories, i.e., Meningioma, Pituitary, and Glioma. In this regards, the paper aims to use  Capsules instead of neurons to build the deep learning architecture. Before introducing the underlying structure of CapsNets and the designed architecture, we briefly review basics of CNNs, which are the common form of deep networks for such classification tasks, and highlight their potential drawbacks which have led to the introduction of Capsule networks.

\subsection{Convolutional Neural Networks} \label{sec:conv}
In brief, CNNs make use of the following three properties: First, units in each layer receive inputs from the previous units which are located in a small neighborhood. This way, elementary features such as edges and corners can be extracted. Then these features will be combined in next layers to detect higher order features. Second important property is the concept of shared weights, which means similar feature detectors are used for the entire image. Finally, CNNs usually have several sub-sampling layers. These layers are based on the fact that the precise location of the features are not only beneficial, but also harmful, because this information tends to vary for different instances~\cite{Lecun:1998}.

Although CNNs have been proved to be useful in many areas, they have several drawbacks specially related to the sub-sampling layers, because these layers give a small amount of translational in-variance and they loose the exact location of the most active feature detectors. Due to the aforementioned reasons, recently a new architecture called Capsule networks is introduced~\cite{Hinton:2017}, which is more robust to translation and rotation. These networks are described in the nest section.

\section{CapsNets for Brain Tumor Classification}\label{sec:WTE}
\begin{figure*}[ht!]
\centering
\includegraphics[width=0.7\textwidth]{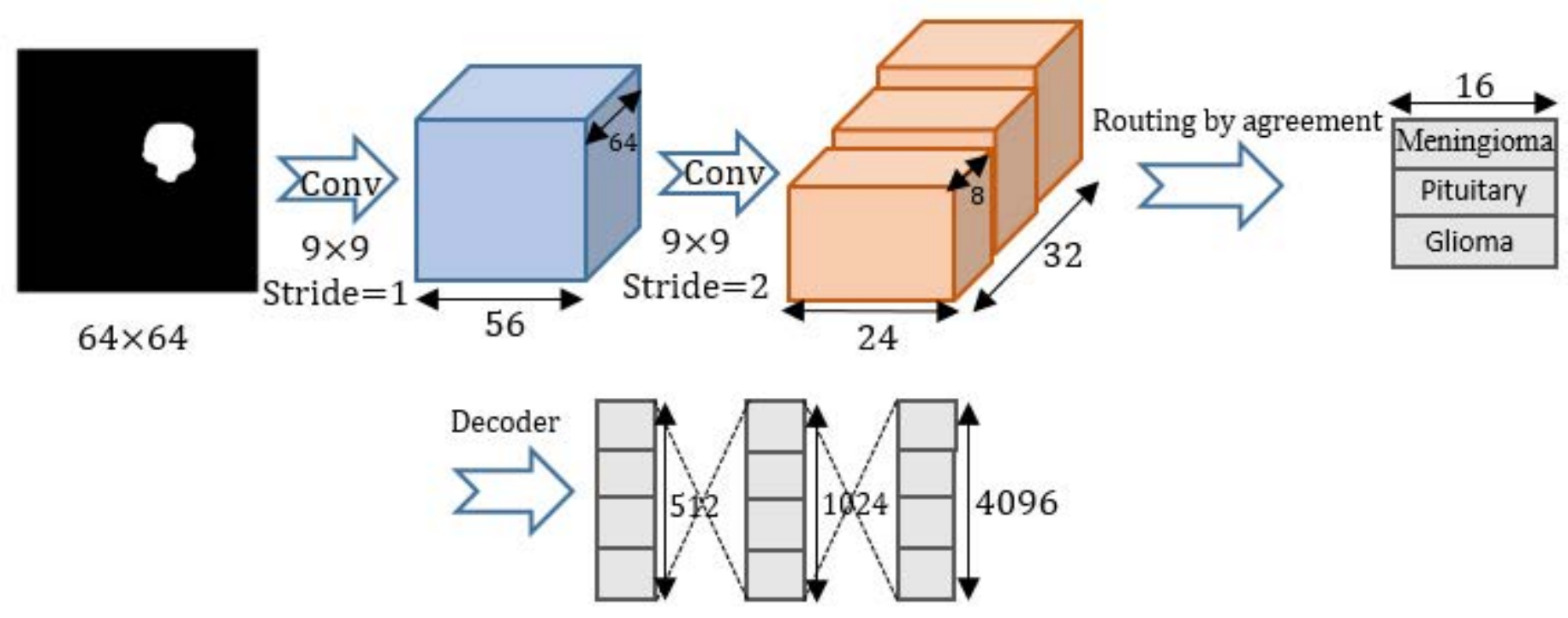}
\caption{\footnotesize Proposed model architecture for brain tumor classification.}
\label{fig:model}
\vspace{-.2in}
\end{figure*}
As stated previously, the goal of the paper is to investigate and design a CapsNet architecture capable of classifying brain tumors as accurately as possible. First, we present the CapsNet properties.

\subsection{Capsule Networks} \label{sec:caps}
Capsules are groups of neurons such that  the activity vectors of these neurons represent various pose parameters and the length of these vectors show the probability that a specific entity exists. The shortcomings of CNNs are mostly related to the pooling layers. As a result, in Capsule networks, these layers are replaced with a more appropriate criteria called ``routing by agreement.'' Based on this criteria, outputs are sent to all parent capsules in the next layer, however, their coupling coefficients are not the same. Each Capsule tries to predict the output of the parent Capsules, and if this prediction conforms to the actual output of the parent Capsule, the coupling coefficient between these two capsules increases. Considering $\u_i$ as the output of capsule $i$, its prediction for parent capsule $j$ is computed~as
\begin{equation}
\hat{\u}_{j|i}=\W_{ij}u_i,
\end{equation}
where $\hat{\u}_{j|i}$ is the prediction vector of the output of the $j^{\text{th}}$ Capsule in a higher level computed by Capsule $i$ in the layer below, and $\W_{ij}$ is the weighting matrix that needs to be learned in the backward pass. Based on the degree of conformation between the capsules in the layer bellow and the parent capsules, coupling coefficients $c_{ij}$ are calculated using the following \textit{softmax} function
\begin{equation}\label{eq:coef}
c_{ij}=\frac{\exp(b_{ij})}{\sum_k\exp(b_{ik})},
\end{equation}
where $b_{ij}$ is the log probability that whether capsule $i$ should be coupled with capsule $j$ and it is initially set to $0$ at the beginning of the routing by agreement process. Therefore, the input vector to the parent capsule $j$ is calculated as follows
\begin{equation}\label{eq:Sj}
\s_j=\sum_ic_{ij}\hat{\u}_{j|i}.
\end{equation}
Finally, the following non-linear squashing function is used to prevent the output vectors of Capsules from exceeding one and forming the final output of each Capsule based on its initial vector value defined in Eq.~\eqref{eq:Sj}
\begin{equation}
\v_j=\frac{\|\s_j\|^2}{1+\|\s_j\|^2}\frac{\s_j}{\|\s_j\|},
\end{equation}
where $\s_j$ is the input vector to Capsule $j$ and $\v_j$ is the output. The log probabilities should be updated in the routing process based on the agreement between $\v_j$ and $\hat{\u}_{j|i}$ using the fact that if the two vectors agree, they will have a large inner product. Therefore,  agreement $a_{ij}$ for updating log probabilities and coupling coefficients is calculated~as follows
\begin{equation}
a_{ij} = \v_j.\hat{\u}_{j|i}.
\end{equation}
Each capsule $k$ in the last layer is associated with a loss function $l_k$, which puts high loss value on capsules with long output instantiation parameters when the entity does not actually exists. The loss function $l_k$ is computed as follows
\begin{equation}
\label{eq:margin}
l_k=T_k \max(0,m^+-||v_k||)^2+\lambda(1-T_k) \max(0,||v_k||-m^-)^2,
\end{equation}
where $T_k$ is $1$ whenever class $k$ is actually present, and is $0$ otherwise. Terms $m^+$, $m^-$, and $\lambda$ are hyper parameters to be indicated before the learning process. The original capsule network architecture presented in~\cite{Hinton:2017} consists of one layer of convolutional filters and two layers of capsules. It has also three layers of fully connected neurons which try to reconstruct the input using the instantiation parameters from the capsule associated with the true label.

\vspace{-.1in}
\subsection{Designed CapsNet} \label{sec:caps}
After exploring several potential architectures, which will be compared later in Section~\ref{sec:EXP} with their associated accuracy, we considered using a model which is similar in nature to the one presented in the original paper, except that it has $64$ feature maps in the convolutional layer instead of 256 as is the case in the original architecture. The summary of the layers of our proposed model (illustrated in Fig.~\ref{fig:model}) is as follows:
\begin{itemize}
\item Inputs to the model are MRI images which are down-sampled to $64\times64$ from $512\times512$, in order to reduce the number of parameters in the model and decrease the training time.
\item Second layer is a convolutional layer with $64\times9\times9$ filters and stride of $1$ which leads to $64$ feature maps of size $56\times56$.
\item The second layer is a Primary Capsule layer resulting from $256\times9\times9$ convolutions with strides of $2$. This layer consists of $32$ ``Component Capsules'' with dimension of $8$  each of which has feature maps of size $24\times24$ (i.e., each Component Capsule contains $24\times24$ localized individual Capsules).
\item Final capsule layer includes $3$ capsules, referred to as ``Class Capsules,' 'one for each type of candidate brain tumor. The dimension of these capsules is $16$.
\item The decoder part is composed of three fully connected layers having $512$, $1024$ and $4096$ neurons, respectively. We note that, the number of neurons in the last fully connected layer is the same as the number of pixels in the input image, as the goal is to minimize the sum of squared differences between input images and reconstructed ones.
\end{itemize}
One of the problems that is observed using CapsNets for the problem at hand with several parameters to be learned and relatively small-scale dataset, is over-fitting. In our different test experiments, we observed that the performance of the trained CapsNet based on the above mentioned specifications was high for training data, but degraded  noticeably on the test data. In other words, careful care is required in the training stage to have a reasonable generalization capability. We adopted early-stopping~\cite{Goodfellow:2016} approach to overcome this problem. According to this approach, at the end of each epoch in the training process, model is tested on a validation set, and training continues to the point that validation accuracy starts to decrease.

For the goal of tumor type classification, two types of images can be used as the input to the aforementioned Capsule network. We can use either the whole brain tissue as the input, or instead, the tumor regions can be segmented first and then use these regions as the input to the classification model. As stated in the CapsNet original paper, Capsules tend to model everything in the input image, thus they do not perform as good as possible for images with miscellaneous backgrounds. Due to this fact, we expect our Capsule network to have a better result when fed with segmented tumors instead of the whole brain images. This is further explored next in Section~\ref{sec:EXP}.

\section{Experimental setup} \label{sec:EXP}
\begin{table}[t!]
 \centering
 \caption{\footnotesize Brain tumor classification accuracy based on different Capsule network architectures.}
 \vspace{.1in}
 \label{tab:arch}
\begin{tabular}{ |l|c| }
\hline
\textbf{Capsule Network Architecture} & \textbf{Prediction Accuracy}\\
\hline
Original architecture & 82.30\%\\
\hline
Two convolutional layers with & \\ 64 feature maps each & 81.97\%\\
\hline
One convolutional layer with &\\ 64 feature maps & \textbf{86.56}\%\\
\hline
One convolutional layer with&\\ 64 feature maps &83.61\%\\ and 16 primary capsules & \\
\hline
One convolutional layer with&\\ 64 feature maps and&82.30\%\\  32 primary capsules of dimension 4 & \\
\hline
Three fully connected layers  &\\ with 1024, 2048&83.93\%\\ and 4096 neurons & \\
\hline
\end{tabular}
\vspace{-.2in}
\end{table}

To test our proposed approach, we have used the data set presented in References~\cite{Cheng:2015,Cheng:2016}. This data set contains $3,064$ MRI images of $233$ patients diagnosed with one of the aforementioned three brain tumor types. The most important property of this data set is that it includes both the brain images and the segmented tumors, which enables us to perform experiments on both types of inputs.

\begin{figure}[t!]
\centering
\includegraphics[width=0.35\textwidth]{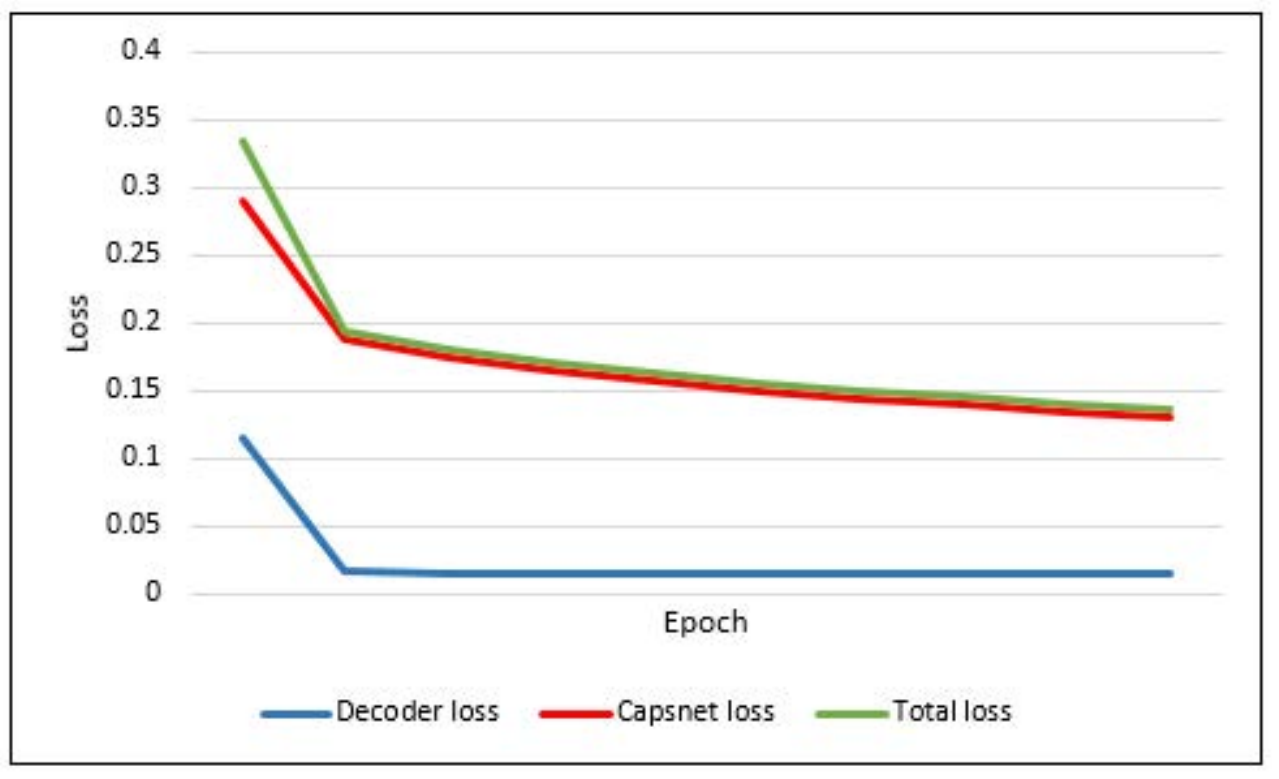}
\caption{\footnotesize Loss values of each of the 10 epochs for the proposed architecture.}
\label{fig:plot}
\vspace{-.1in}
\end{figure}

The first part of our experiments is allocated to testing different kinds of Capsule network architectures. We have changed different components of the original framework and calculated the prediction accuracy. Table~\ref{tab:arch} outlines the obtained results. According to these results, reducing the number of feature maps from $256$ (as is the case in the original architecture) to $64$ leads to the highest accuracy. However, there are many more architectures that can be explored, which is the focus of our ongoing research work.
Next, we evaluate the total loss in a Capsule network, which is composed of two parts: CapsNet loss and Decoder loss. The former calculates the miss-classification error and is determined using Eq.~\eqref{eq:margin}. The latter is related to the reconstruction part and is calculated using the square error between the input and the reconstructed image. This loss contributes to the total loss with a smaller weight. We have trained our proposed architecture for $10$ epochs and computed the three losses at the end of each epoch as shown in Fig.~\ref{fig:plot}. It is observed that training is faster at the beginning and the total loss is mostly dependent on the CapsNet~loss.

\begin{figure}[t!]
\centering
\includegraphics[width=0.35\textwidth]{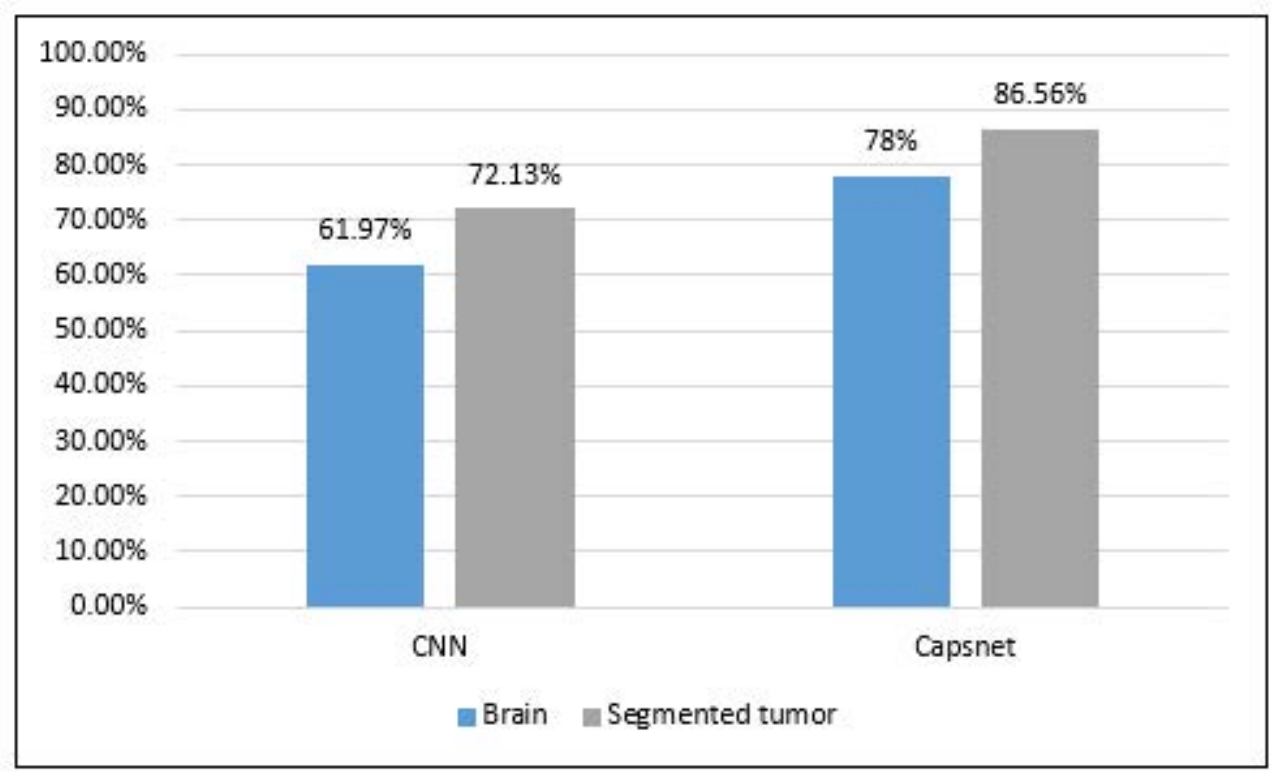}
\caption{\footnotesize CapsNet and CNN accuracy for brain and segmented tumors images.}
\label{fig:cnn}
\vspace{-.2in}
\end{figure}

After selecting the best architecture for the Capsule network, we have compared its classification accuracy with a conventional CNN over the same dataset. The CNN used for comparison is adopted from~\cite{Justin:2017}, which has investigated the problem of brain tumor classification on the same data set used in this paper. The layers of this CNN are constructed as follows:
\begin{itemize}
\item Convolutional layer with $64\times5\times5$ filters and strides of $1$.
\item $2\times2$ Max-Pooling.
\item Convolutional layer with $64\times5\times5$ filters and strides of $1$.
\item $2\times2$ Max-Pooling.
\item Fully connected layer of $800$ neurons.
\item Fully connected layer of $800$ neurons.
\item Fully connected layer of $3$ neurons.
\end{itemize}
We compared Capsule network with the CNN for both brain images and segmented tumors. Fig.~\ref{fig:cnn} illustrates the comparison results. Based on these results, it is observed that CapsNet outperforms CNN for both types of inputs. As it is stated previously, Capsules tend to account for everything in the input image even in the background, and considering the fact that brain MRI images are taken from different angles such as Sagital and Coronial, backgrounds have lots of variations. Therefore, CapsNet can not handle brain images as good as segmented tumor images, and this may be one of the reasons for the CapsNet architectures to provide lower accuracy for brain images in comparison to the case where segmented tumors are used as the input.
Nevertheless, CapsNet's result is superior to that of the CNN for brain tumor classification, which shows Capsule networks' advantageous over CNNs. One reason behind success of CapsNets in providing better brain tumor classification results can be attributed to the fact that CapsNets can handle data sets with smaller number of samples better than CNNs.

\begin{figure}[t!]
\centering
\includegraphics[width=0.22\textwidth]{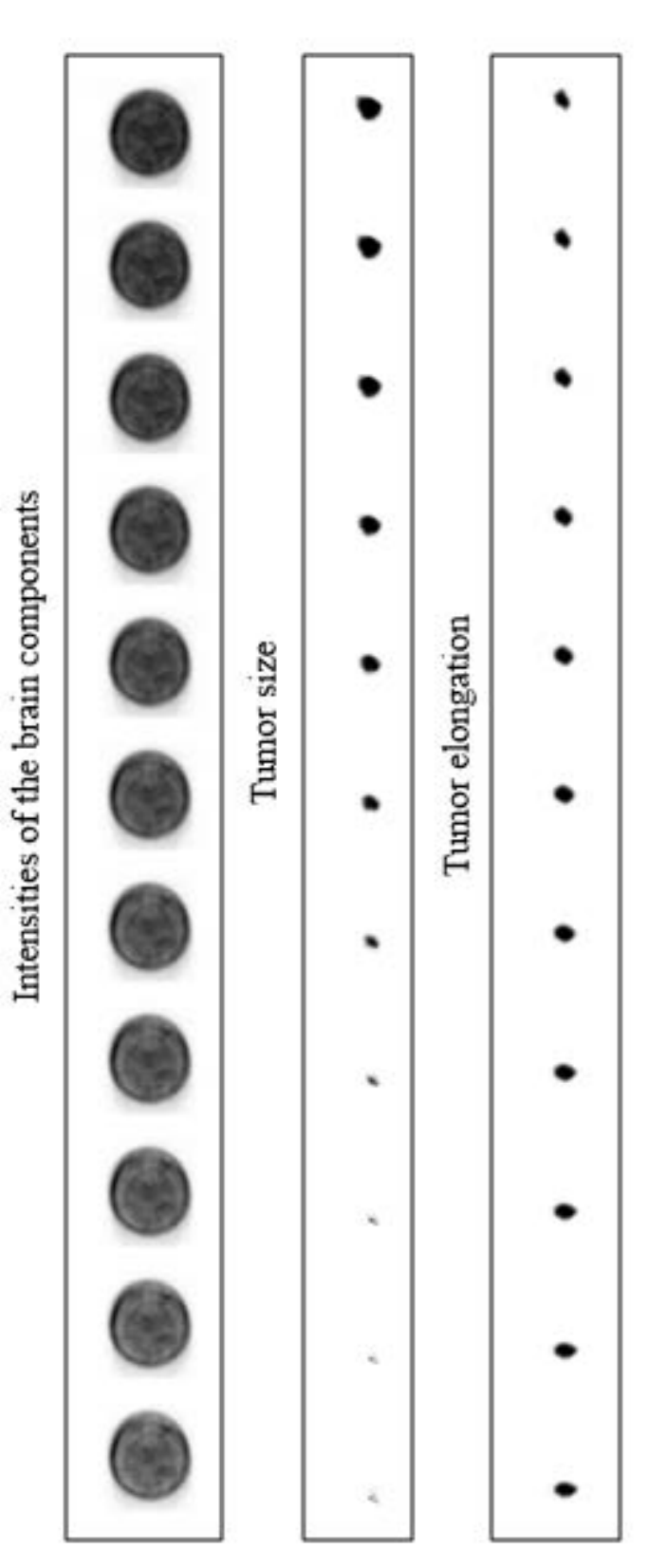}
\caption{\footnotesize Features detected by tweaking activation vectors of Class Capsules.}
\label{fig:tweak}
\vspace{-.2in}
\end{figure}

Finally, we investigate the output of the last layer in the CapsNet (referred to as the ClassCap), which is a vector containing the pose features, however, the CapsNet determines what features to learn on it's own. To provide better explainability, therefore,  one option is to tweak them and try to reconstruct the input image using these tweaked vectors (by tweaking we refer to adding small numbers to the original vectors). When we visualized these tweaked vectors, one can identify/capture the nature of learned features. Fig.~\ref{fig:tweak} illustrates some of the results together with what type of features they seem to be related to. Each column represents one particular reconstructed input using the tweaked features. For instance, the special feature learned in the second column seems to represent the size of tumor as tweaking this feature has changed the size. Similarly, the third column seems to be related to how wide the tumor is.
\vspace{-.1in}
\section{Conclusion}  \label{sec:con}
In this work, we have investigated the use of newly proposed Capsule networks for the problem of brain tumor type  classification. Since these networks can handle small number of training samples, and units in these networks are equivariant, they outperform CNNs in tumor classification problem. We could also increase the accuracy by changing the number of feature maps in the convolutional layer of the Capsule network. Furthermore, based on our experiments, these networks can perform better for the segmented tumors than for the whole brain images.
In future, we plan to investigate effects of having more capsule layers on the classification accuracy. 


\end{document}